%% file: main_preprint.tex
\documentclass[11pt, a4paper, logo, copyright]{preprint-asset/kunumi}

\usepackage[authoryear, sort&compress, round]{natbib}
\usepackage[]{mdframed}
\usepackage{anyfontsize}
\usepackage{listings}
\usepackage{hyperref}
\usepackage{url}
\usepackage{graphicx}
\usepackage{mathrsfs} %
\usepackage{etoolbox}
\usepackage{cleveref}
\usepackage{tcolorbox}
\usepackage{tabularx}
\usepackage{colortbl}
\usepackage{booktabs}       %
\usepackage{amsfonts}       %
\usepackage{nicefrac}       %
\usepackage{microtype}      %
\usepackage{subcaption}
\usepackage{multirow}
\usepackage{subcaption}
\usepackage{ragged2e}
\usepackage{enumitem}%
\setlist[itemize]{noitemsep, topsep=0pt}
\usepackage{enumitem,kantlipsum}
\usepackage{tabularx}
\usepackage{ragged2e} %
\usepackage{booktabs} %
\usepackage{xr} 
\usepackage{xcolor}

\newlength\savewidth

\definecolor{baselinecolor}{HTML}{d6eaf8}

\definecolor{mygray}{gray}{0.4}

\definecolor{darkgreen}{rgb}{0, 0.5, 0}

\AtBeginEnvironment{tcolorbox}{\tiny}
\usepackage[utf8]{inputenc} 
\usepackage[T1]{fontenc}    
\usepackage{hyperref}       
\usepackage{url}            
\usepackage{booktabs}       
\usepackage{amsfonts}       
\usepackage{nicefrac}       
\usepackage{microtype}      
\usepackage{xcolor}         
\usepackage{amsmath}
\usepackage{amsthm}

\usepackage{mathtools}
\usepackage{algorithm}
\usepackage{algorithmicx}
\usepackage{booktabs}
\usepackage{multirow}
\usepackage{tabularx}
\usepackage{enumitem}
\usepackage{tikz}
\usepackage{amsmath}
\usepackage{xcolor}
\usepackage{nameref}
\usepackage[dvipsnames]{xcolor}
\usetikzlibrary{arrows.meta, positioning, fit, backgrounds, shapes.geometric, calc, decorations.pathreplacing}
\usepackage{siunitx}
\usepackage{graphicx,subcaption}
\usepackage{algorithm}
\usepackage{algpseudocode}
\usepackage{float}
\usepackage{subcaption}

\usepackage{xspace}

\usepackage{hyperref} 

\usepackage{longtable}  

\begin{document}
\title{Performance Evaluation of GraphCast for Medium-Range Weather Forecasting over Brazil}
%
%

%

\author[1]{Wolfgang R. Rowell Jr.}
\author[1]{Lucas S. Kupssinskü}
\affil[1]{MALTA, Machine Learning Theory and Applications Lab, PUCRS, Porto Alegre, Brazil}

%

\input{sections/abstract}


\maketitle              

\input{sections/introduction}

\input{sections/background}

\input{sections/methodology}

\input{sections/results}

\input{sections/conclusions}

\bibliography{sample-base}

\input{sections/appendix}

\end{document}

%% file: sections/abstract.tex
\begin{abstract}
The paradigm of global weather forecasting is rapidly shifting with the emergence of Machine Learning Weather Prediction models (MLWP). While these data-driven architectures demonstrate remarkable global skill, regional benchmarks in the Global South remain scarce, leaving their efficacy in complex, highly convective environments largely unverified. This study evaluates the performance of GraphCast operational against the deterministic ECMWF IFS HRES as baseline across four distinct Brazilian climatic sub-regions.  Utilizing a scalable, cloud-native pipeline and the WeatherBench-X framework for benchmarking weather models, we assess selected tropospheric variables ($T_{850}$, $Q_{850}$, $Z_{500}$) over four selected seasonal windows, employing the operational IFS analysis as the ground truth to calculate the statistical metrics for both models. Results reveal a regime-dependent skill profile. During the austral winter, GraphCast underperforms in the medium range (lead days 2-7) for $Z_{500}$ when resolving fast-propagating baroclinic systems over southern Brazil, but regains an advantage in the extended range, where its inherent smoothing of chaotic small-scale variability becomes beneficial under deterministic skill metrics. Conversely, during the austral summer wet season, GraphCast accurately captures large-scale moisture transport while intrinsically dampening the high-frequency convective variability that degrades deterministic NWP temperature forecasts. These findings establish a baseline for Brazil and define the specific physical boundaries that will guide future ``tropicalization'' efforts, aiming to optimize these foundational AI models for regional resilience.
\end{abstract}

%% file: sections/introduction.tex
\section{\label{sec:secao1}Introduction}

Numerical Weather Prediction (NWP) has been the operational foundation of medium-range forecasting for decades, providing the deterministic guidance on which a wide range of weather-sensitive sectors---energy, agribusiness, transportation, insurance, and civil defense---rely to manage risk~\cite{Lynch2008,Coiffier2011-jz}. Despite continuous improvement, the physics-based paradigm faces well-documented limitations in computational cost, the parameterization of subgrid processes, and the chaotic
predictability horizon~\cite{ECMWF-Chaos}, with these constraints further amplified by the non-stationarity of a warming climate~\cite{Slater2023}.

A recent generation of data-driven Machine Learning Weather Prediction (MLWP) models has reframed this trade-off. By learning the non-linear dynamics of the atmosphere directly from reanalysis data~\cite{ECMWF-ERA5}, models such as Google DeepMind's GraphCast match or exceed the skill of the ECMWF Integrated Forecasting System deterministic model (IFS HRES) on global benchmarks at a fraction of the inference cost~\cite{Lam2023,BenBouallgue2024,Price2024}. The reported skill, however, is established on \emph{globally averaged} metrics, leaving a question that is operationally critical but largely unanswered in the peer-reviewed literature: does this skill generalize to specific, climatically complex regions?

Assuming that a generalization gap exists, its implications may be particularly significant in Brazil. Spanning the equatorial Amazon, the semi-arid Northeast, the convective Cerrado of the central plateau, and the temperate South~\cite{Nimer1979}. The country exposes any forecasting system to a wide range of atmospheric regimes, from tropical dynamics governed by moist thermodynamics to fast-propagating extratropical baroclinic systems. While the regional verification of traditional NWP over South America has recently been advanced~\cite{ECMWF2025}, an analogous, regime-stratified assessment of operational MLWP models remains absent. Resolving this question is more than an academic exercise: understanding the extent to which forecasting systems generalize across diverse climatic regimes has important implications for weather-sensitive sectors worldwide. This is especially relevant in the Global South, where regions often combine high climate variability with strong socioeconomic dependence on weather and climate conditions. Improved forecasting skill can support renewable energy operations~\cite{Street2025}, agricultural planning, transportation and logistics, water-resource management, and the assessment of climate-related risks.

To address this gap, we present a regional benchmark of GraphCast against the IFS HRES baseline over four Brazilian climatic sub-regions. Both forecasts are verified against the operational IFS analysis as ground truth, avoiding the circular bias of evaluating a data-driven model against its own training distribution. Forecast skill is quantified
through the Root Mean Square Error (RMSE), the Anomaly Correlation Coefficient (ACC), and the normalized RMSE difference for three tropospheric variables evaluated across four seasonal windows: temperature ($T_{850}$) and specific humidity ($Q_{850}$) at 850~hPa, which characterize lower-tropospheric thermodynamics and moisture transport, and geopotential ($Z_{500}$) at 500~hPa, the standard proxy for mid-tropospheric
synoptic-scale circulation.

The contributions of this paper are threefold: \textbf{(i)} the first regime-stratified, sub-regional benchmark of an operational MLWP model over Brazil; \textbf{(ii)} a reproducible, cloud-native evaluation pipeline built on WeatherBench-X that harmonizes ECMWF Open Data and Google DeepMind WeatherNext outputs into a common analytical substrate; and \textbf{(iii)} a diagnostic mapping between Brazilian climatic regimes and the conditions under which the inductive paradigm gains or loses
skill relative to the deterministic baseline, establishing operational boundaries that motivate the future regional ``tropicalization'' of foundational AI weather models.

%% file: sections/background.tex
\section{Background}
The Earth's atmosphere is a highly complex, chaotic dynamical system driven by continuous thermodynamic interactions and fluid motion~\cite{wallace2006atmospheric}. Historically, the scientific community has approached the monumental task of predicting its future states through a strictly deterministic and physics-based lens. By conceptualizing the atmosphere as a fluid envelope governed by fundamental conservation laws—such as the conservation of mass, momentum, and energy—meteorologists have sought to translate natural phenomena into rigorous mathematical frameworks. This traditional paradigm, which relies on deductive inference to simulate atmospheric evolution forward in time from a known initial state, forms the conceptual and operational foundation of NWP.

\subsection{Numerical Weather Prediction and the Deterministic Paradigm}
To understand the shift brought by artificial intelligence, one must first examine the established baseline: the Numerical Weather Prediction. At its core, NWP treats weather forecasting as a massive, deterministic initial value problem. The atmosphere is modeled as a fluid envelope governed by the fundamental laws of physics specifically, the conservation of momentum (the Navier-Stokes equations), mass, energy, and water vapor. For global operational models like the ECMWF IFS, these laws are expressed through a set of highly complex, non-linear partial differential equations known as the primitive equations~\cite{Coiffier2011-jz}. Because these equations cannot be solved analytically, they require the immense computational power of supercomputers to iteratively simulate the atmosphere's evolution forward in time.

To solve these continuous equations digitally, the Earth's atmosphere must be mapped onto a discrete, three-dimensional computational grid. However, this discretization introduces a fundamental limitation: finite resolution~\cite{Coiffier2011-jz}. Atmospheric phenomena that occur at scales smaller than the model's grid boxes such as cloud microphysics, turbulence, and localized tropical convection cannot be explicitly simulated. Instead, their large-scale effects must be mathematically approximated through a process called parameterization~\cite{Coiffier2011-jz}. In regions like Brazil, where intense subgrid scale convective processes drive a significant portion of the weather, parameterization remains a persistent source of model uncertainty and forecast error~\cite{ECMWF2025}.

Within this established paradigm, the ECMWF IFS is widely recognized as the global benchmark. Rather than maintaining a patchwork of different models, the IFS functions as a unified "software engine" that couples atmospheric, oceanic, and land-surface processes. What makes the IFS especially powerful is its configurability: the same underlying core science is used to generate the historical ERA5 reanalysis (which underpins climate monitoring and MLWP training) as well as the deterministic High-Resolution forecast. Operating at a 9 km spatial grid to provide a single, sharp picture out to ten days, the IFS HRES serves as the definitive standard against which emerging artificial intelligence models must be rigorously quantified~\cite{BenBouallgue2024}.

Before the IFS HRES can calculate the future, however, it requires a mathematically precise snapshot of the present. At the start of every 6-hour cycle, this initial state is generated through data assimilation, a sophisticated statistical step that optimally blends a short-term background forecast with millions of real-world observations from the Global Observing System (GOS)~\cite{WMO-GOS}. The output of this blending process is known as the analysis, the most accurate possible numerical representation of the current Earth system state~\cite{Coiffier2011-jz}.

In the context of this study, the operational ECMWF IFS analysis, is known as the 0th frame (or the weather data at $t=0$), and serves as the absolute ground truth~\cite{rasp2024weatherbench2}. By utilizing this operational analysis rather than historical reanalysis datasets such as ERA5, we establish a rigorous, unbiased benchmark. This allows for a fair evaluation of the operational skill of the data-driven GraphCast operational model, that was trained with ERA5 and fine-tuned with IFS analysis data, directly against the state-of-the-art deterministic IFS HRES baseline~\cite{Lam2023}\cite{rasp2024weatherbench2}.

\subsection{The Shift to Machine Learning Weather Prediction (MLWP)}
In contrast to the physics-based NWP approach, the recent emergence of Machine Learning Weather Prediction introduces a fundamental paradigm shift in how the atmosphere is modeled. At a fundamental level, an ML-based prediction is the result of an inductive rather than a deductive inference. Instead of explicitly solving the primitive equations, these data-driven models utilize deep neural networks to learn the complex, non-linear spatiotemporal dynamics of the Earth system directly from decades of historical weather data. This paradigm shift in terms of logic implies that a forecast becomes a plausible outcome given what has been learned from previous data. The primary operational advantage of this approach is that, once trained, the inference process—generating the actual forecast bypasses the traditional supercomputing bottlenecks of NWP, delivering predictions at a speed several orders of magnitude faster and at a much lower computational cost~\cite{BenBouallgue2024}.

The journey of data-driven forecasting from an exploratory academic concept to a leading operational methodology has been remarkably fast. Initial efforts in the late 2010s utilized simple neural networks operating at coarse resolutions. While these first-generation models were crucial for establishing the field and prompting the creation of reproducible evaluation frameworks like WeatherBench in 2020~\cite{Weatherbench-2}, their forecast skill remained considerably less accurate than contemporary physical models. However, the pace of progress accelerated dramatically in 2022, marking a clear turning point characterized by rapid architectural evolution and significant increases in spatial resolution. Multiple AI models began to progressively approach the performance of state-of-the-art numerical systems. Among these breakthroughs, Google DeepMind's GraphCast emerged as one of the most incredible achievements in the field, demonstrating the unprecedented capability to consistently surpass the deterministic ECMWF IFS HRES baseline across numerous atmospheric variables and pressure levels~\cite{Lam2023}. This remarkable leap in predictive skill transitions the conversation from whether ML models can predict the weather, to understanding the specific internal mechanisms that allow it to do so with such high fidelity.

\subsection{GraphCast Model and Architecture}
While earlier MLWP attempts relied heavily on standard Convolutional Neural Networks (CNNs), representing the Earth's atmosphere on a standard latitude-longitude grid introduces severe geometric distortions, particularly converging at the poles~\cite{Weyn2020}. To overcome this spatial bias, Google DeepMind's GraphCast, following the impressive results obtained by Keisler months before~\cite{Keisler2022},  utilized an upgraded Graph Neural Network (GNN) architecture. By representing atmospheric states as nodes and spatial relationships as edges, GNNs can model the spherical nature of the globe seamlessly, facilitating the highly efficient transfer of meteorological information across vast distances.

Internally, GraphCast operates through a highly optimized "Encode-Process-Decode" framework. The model does not perform its heavy computations directly on the native latitude-longitude input grid. Instead, the Encoder maps the high-resolution grid data onto a multi-scale icosahedral mesh, a nearly uniform polyhedron that encapsulates the globe. Once the data is embedded onto this mesh, the processor executes a series of message passing steps. Because the mesh contains multi-scale connections (short edges for local interactions and long edges for global teleconnections), the processor can efficiently resolve both small-scale localized convection and massive synoptic-scale planetary waves with minimal computational overhead. Finally, the decoder maps these updated node features back from the icosahedral mesh to the standard latitude-longitude grid to output the final physical variables. 

Temporally, GraphCast functions as an autoregressive model. To forecast the weather, it requires two initial atmospheric states as input: the current state ($t=0$) and the state six hours prior ($t = -6h$). Using these two frames to establish the atmospheric momentum and trajectory, the model predicts the state of the atmosphere a single 6-hour step into the future ($t = +6h$). To generate medium-range forecasts up to 10 days, GraphCast feeds its own predictions back into itself autoregressively. This capacity to capture complex, long-range spatial dependencies while maintaining computational efficiency at inference time is what allows the model to rival and often surpass the deterministic operational state-of-the-art baseline.

\subsection{Climatic sub-regions of Brazil}
\begin{figure}[H]
  \centering
  \includegraphics[width=0.7\linewidth]{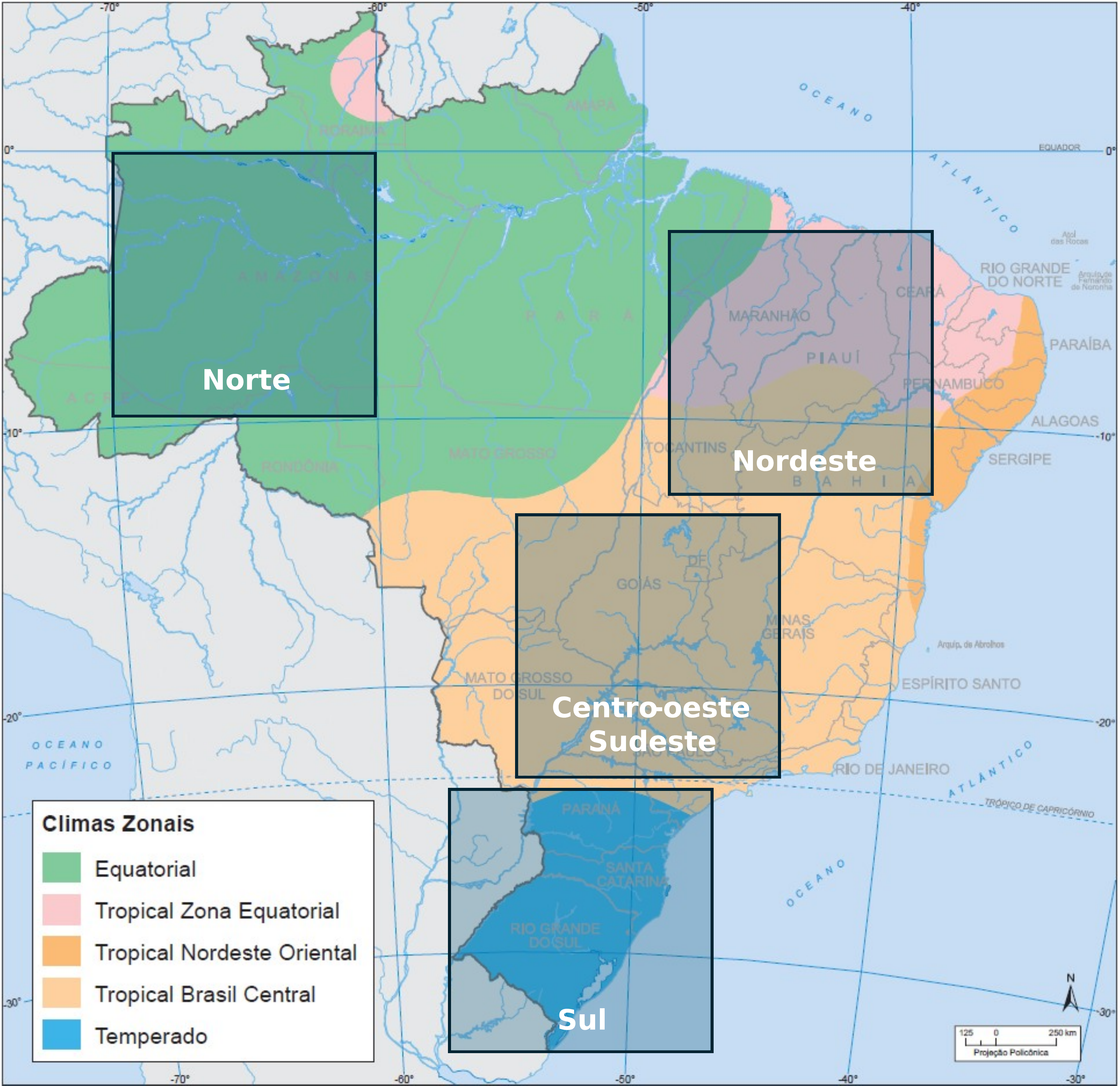}
\caption{\label{fig:fig10}Climatic sub-regions of Brazil and the selected areas for analysis: North, Northeast, Central-West/Southeast, and South. Legend: Equatorial (green), Tropical Equatorial (pink), Tropical Oriental (brown), Tropical Central (yellow) and Subtropical/Temperate (blue). Source: base map IBGE~\cite{IGBE-ClimaticMap}, adapted by the authors}
\end{figure}

The continental expanse of Brazil and its exposure to a broad range of atmospheric processes result in a remarkable diversity of climates. Although fundamentally a tropical nation with elevated temperatures prevailing year-round, its vast territorial extent is geographically organized into distinct climatic zones. As illustrated in the Figure~\ref{fig:fig10}, each of these regions fosters a unique environment characterized by significant inter-seasonal variability, driven primarily by fluctuations in pluviosity and, at higher latitudes, temperature~\cite{Nimer1979}.

\textbf{Equatorial}
The Equatorial climate, encompassing the majority of the Amazon Basin, is characterized by persistent high temperatures and abundant pluviosity. Its defining feature is a negligible thermal amplitude; temperatures remain uniformly elevated throughout the annual cycle. Consequently, seasonal variation is dictated exclusively by precipitation rather than thermal shifts. While the region is consistently humid, the "dry season"—typically occurring from July to November, contingent upon latitude—manifests as a period of diminished rainfall rather than an absolute drought, sustaining the dense biomass of the Amazon rainforest~\cite{CavalcantiFerreira2021}.

\textbf{Tropical - Equatorial Zone Tropical}
Functioning as a transitional climate, the Tropical - Equatorial Zone is primarily localized along the peripheries of the central Amazon Basin (e.g., portions of Roraima, Amapá, and the eastern Amazon). Its principal divergence from the strictly Equatorial zone is the emergence of a brief but distinct dry season. While retaining the high thermal averages and substantial annual precipitation of the equatorial core, its seasonality is more pronounced. During the austral winter, precipitation decreases significantly, creating ecotonal landscapes where dense rainforest transitions into open forests or savanna enclaves~\cite{CavalcantiFerreira2021}.

\textbf{Tropical - Oriental Northeast Tropical}
Situated along the eastern littoral of Brazil’s Northeast, the Tropical - Oriental Northeast (frequently designated as \textit{Tropical Litorâneo}) is heavily governed by maritime influences and the Atlantic trade winds. This zone presents an anomalous pluviometric regime compared to the national average: precipitation is concentrated during the autumn and austral winter months (March to August), whereas the summer months constitute the driest period of the year. Thermal averages remain consistently high, though moderated by persistent oceanic breezes~\cite{CavalcantiFerreira2021}.

\textbf{Tropical - Central Brazil}
Encompassing the expansive central plateau and the \textit{Cerrado} biome, the Tropical - Central Brazil represents the quintessential tropical savanna climate. Its defining climatological trait is a pronounced seasonality predicated on moisture availability. The annual cycle is sharply divided into a hot, high-precipitation summer (October to April) and a warm, severely arid winter (May to September). During the winter months, relative humidity frequently drops to critical levels, inducing significant physiological stress on the vegetation and drastically altering the landscape until the return of the summer rains~\cite{CavalcantiFerreira2021}.

\textbf{Subtropical / Temperate}
Predominantly located south of the Tropic of Capricorn, this macro-region diverges fundamentally from the rest of the country due to its marked thermal seasonality. It is crucial, however, to distinguish between its two primary regimes: the widespread Subtropical climate and the localized Temperate enclaves. The Subtropical regime characterizes the broader regional extent and lowlands, where precipitation is uniformly distributed throughout the year without a discernible dry season, and summers are marked by high heat and humidity. Conversely, a genuinely Temperate climate exists exclusively within the high-altitude zones—specifically the elevated plateaus, mountainous terrains, and \textit{Serras} spanning the states of Paraná, Santa Catarina, and Rio Grande do Sul~\cite{CavalcantiFerreira2021}.  In these higher elevations, seasonal shifts are strictly driven by sharp temperature variations. Winters present genuinely low temperatures, frequently experiencing widespread frost and occasional snowfall, making these southern highlands the only Brazilian domain that exhibits a recognizable, temperature-driven four-season cycle~\cite{CavalcantiFerreira2021}.

\subsection{Meteorological Variables of Interest}
To evaluate the predictive skill of forecasting models over a domain as geographically complex and expansive as Brazil, it is essential to select meteorological variables that capture both large-scale synoptic forcing and lower-tropospheric thermodynamics. Rather than evaluating highly localized surface variables—which are heavily influenced by subgrid-scale topography, land-cover variations, and immediate diurnal heating cycles—this study focuses on fundamental upper-air variables evaluated at specific pressure levels.

In atmospheric modeling, an isobaric surface is a layer of constant atmospheric pressure, and it is the standard reference frame for analyzing fluid dynamics. Because the atmosphere is a compressible fluid, the geometric height of these surfaces fluctuates dynamically in response to changes in air mass temperature and density. Utilizing constant-pressure surfaces rather than fixed geometric altitudes significantly simplifies the governing thermodynamic equations and provides a more coherent representation of three-dimensional atmospheric flow~\cite{Ahrens2016-ha}. Based on this established framework, our comparative evaluation focuses on three critical tropospheric variables: Temperature at 850 hPa ($T_{850}$), Specific Humidity at 850 hPa ($Q_{850}$), and Geopotential at 500 hPa ($Z_{500}$).

\vspace{1em}
    - \textbf{Temperature at 850 hPa ($T_{850}$):} Situated at approximately 1.500 meters above sea level, the 850 hPa pressure level represents the top of the planetary boundary layer in many regions. $T_{850}$ is utilized to evaluate lower-tropospheric thermal advection and the characteristics of advancing air masses. Because this layer is largely decoupled from the immediate, noisy effects of surface radiative heating and cooling, it provides a highly reliable signal of the true atmospheric thermal state, making it indispensable for predicting heatwaves or polar outbreaks such as the \textit{friagens} in the Amazon and Center-West~\cite{Ahrens2016-ha}.

\vspace{1em}
    - \textbf{Specific Humidity at 850 hPa ($Q_{850}$):} In the tropical and subtropical South American context, low-level moisture transport is the primary fuel for convective weather. $Q_{850}$ tracks the availability and movement of this water vapor. It is fundamentally important for capturing the dynamics of the South American Low-Level Jet (SALLJ), the phenomenon often referred to as "flying rivers" which transports massive quantities of moisture from the Amazon basin down to the central and southern sub-regions. Errors in $Q_{850}$ directly translate to failures in predicting deep convection, precipitation, and cloud cover, which heavily impact the operational predictability of hydro and solar power generation~\cite{Ahrens2016-ha}.

\vspace{1em}    
    - \textbf{Geopotential at 500 hPa ($Z_{500}$):} Located in the mid-troposphere approximately 5.500 meters above sea level, $Z_{500}$ is the standard meteorological proxy for large-scale synoptic circulation. It dictates the steering mechanisms of global weather systems. Accurately forecasting $Z_{500}$ is critical for predicting the progression of transient systems, such as the deep troughs and ridges that guide cold fronts from the extreme south of the continent into the Brazilian subtropics and Southeast regions~\cite{Ahrens2016-ha}.

%% file: sections/methodology.tex
\section{Methodology and Data}
\vspace{1em}
To address the research question of this study: \textit{How does GraphCast perform across Brazil’s diverse climatic sub-regions compared to a state-of-the-art NWP baseline?} we employ a comprehensive and reproducible evaluation framework to assess their operational predictive skill. A critical component of this research is the underlying data engineering required to process, align, and analyze large volumes of atmospheric data without introducing methodological biases. To achieve this, we developed a custom, scalable cloud-based pipeline that aggregates and harmonizes 27 distinct datasets, sourced from ECMWF Open Data~\cite{ECMWF-opendata} and Google Deepmind's WeatherNext, totaling 145 GB of information collected~\cite{Lam2023}. The following subsections detail the geographical and temporal domains selected for this regional benchmark, the architecture of the data ingestion and processing pipeline, and the mathematical formalization of the evaluation metrics calculated utilizing the WeatherBench-X framework.~\cite{Weatherbench-X}.

\subsection{Spatial and Temporal Domain}
\vspace{1em}
\subsubsection{Spatial choice} 
Evaluating atmospheric models over a continent-sized territory requires a methodology that prevents the dilution of regional forecasting errors. When evaluation metrics are calculated, they are spatially averaged over the area of analysis. If this averaging is performed over the entirety of Brazil, compensating errors across vastly different climatic zones can mask specific localized model deficiencies. To ensure a richer, more instructive exploration of the model's capabilities, we subdivided the study area into four distinct bounding boxes of equal geometric dimensions of $10^\circ \times 10^\circ$, covering approximately $1.120 \times 1.120$ km each.

These bounding boxes were designed to represent four broad climatological regions of Brazil—the North, Northeast, a combined Center-West and Southeast region, and the South—as illustrated in Figure~\ref{fig:fig10}. Rather than following the exact boundaries of Brazil's administrative regions, the analysis employs rectangular domains of identical dimensions. This choice ensures that all regional statistics are computed from spatial samples of comparable size, reducing potential biases associated with unequal domain areas and enabling a more consistent comparison of model performance across regions.

The selected domains were positioned to capture the dominant atmospheric characteristics of each macro-region while minimizing overlap with neighboring climatic regimes. This spatial subdivision allows the evaluation framework to isolate regional forecast skill without blending distinct synoptic and thermodynamic environments. The narrow eastern coastal strip of the Northeast was deliberately excluded from the bounding boxes because its highly localized coastal weather regime operates on a different spatial scale than the broader continental mass-field dynamics under evaluation. Including this area would introduce variability that is not representative of the regional-scale processes targeted in this study.

\subsubsection{Temporal choice} 
To evaluate the models under varying atmospheric circulation patterns, the temporal domain was specifically selected to capture distinct seasonal peaks and transition periods. The benchmark is conducted over four months: March 2024 (Summer-to-Autumn transition), July 2024 (Winter peak), October 2024 (Spring transition), and January 2025 (Summer peak). Evaluating across these specific temporal windows ensures that the data-driven models are tested against the full seasonal spectrum of the Brazilian weather dynamics.

\subsection{Datasets and Data Pipeline}
\begin{figure}[H]
    \centering
    \includegraphics[width=0.85\linewidth]{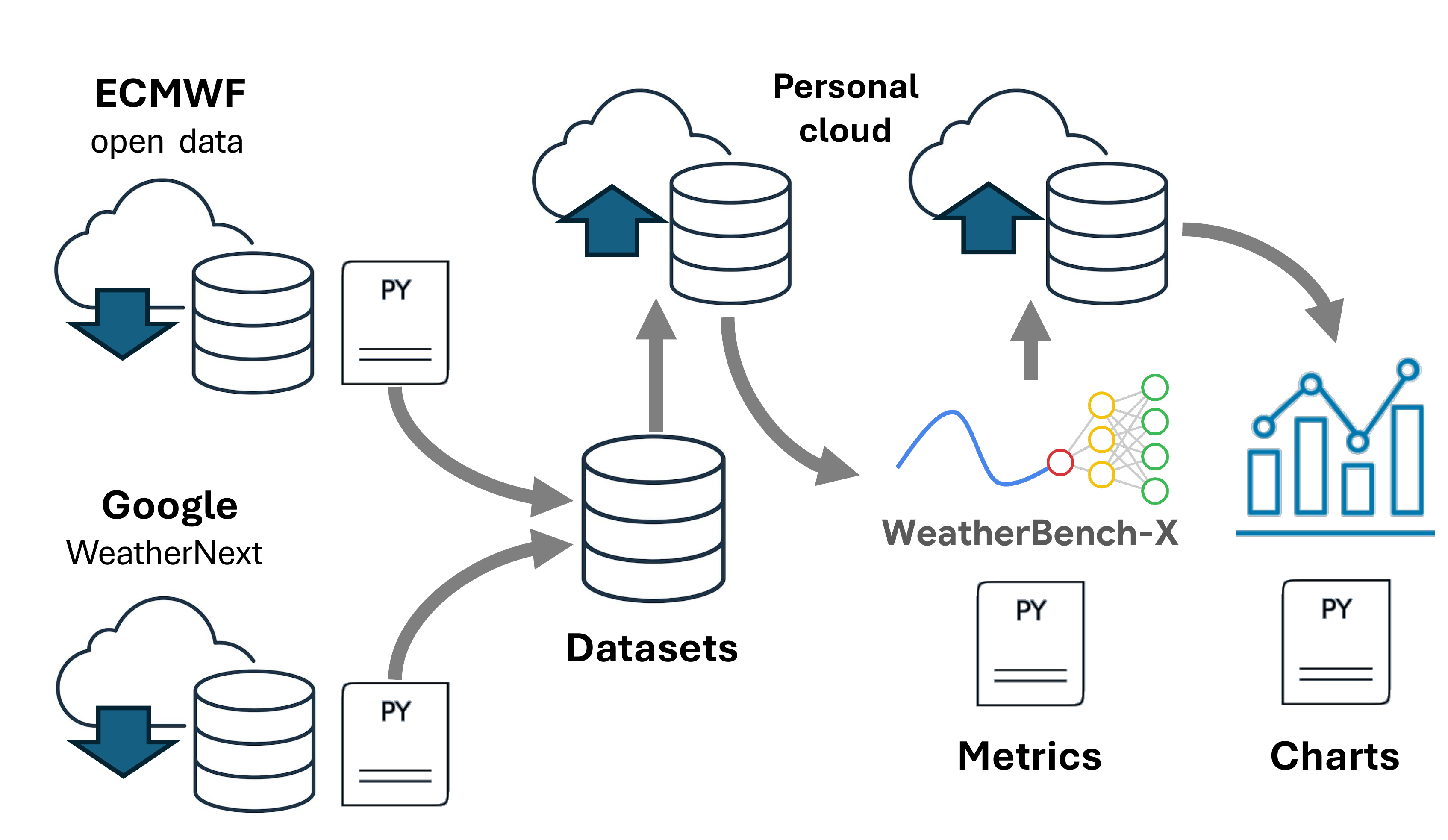}
    \caption{\label{fig:fig3}Pipeline developed to build and use the necessary data. Source: authors.}
\end{figure}

\vspace{1em}
To ensure a seamless and reproducible evaluation, we architect a custom, cloud-native data ingestion and consolidation pipeline. This pipeline extracts the pre-processed forecast data, aligns the dimensional coordinates, and consolidates the outputs into cloud-optimized formats for high-performance evaluation. The overall architecture of this custom data ingestion, harmonization, and evaluation pipeline is illustrated in Figure~\ref{fig:fig3}. This schematic outlines the end-to-end data flow, from the initial extraction of forecasts from disparate cloud repositories to the final generation of consolidated ZARR datasets and analytical visuals.

\subsubsection{Data Sourcing and Extraction}

To ensure a consistent comparison framework, all forecasts were evaluated on the native spatial and temporal grids provided by their respective operational data services. Although the original IFS HRES model is produced at approximately $0.1^\circ$ resolution, the ECMWF Open Data service distributes forecasts on a standardized $0.25^\circ \times 0.25^\circ$ grid with 6-hour temporal intervals. This matches the operational GraphCast data format, as well as the ERA5 reanalysis used as reference, eliminating the need for additional spatial or temporal interpolation.

GraphCast operational forecasts are available for four daily initialization times (00Z, 06Z, 12Z, and 18Z) with lead times extending from $+6$ h to $+240$ h. In contrast, IFS HRES forecasts initialized at 00Z and 12Z extend to $+240$ h, whereas the 06Z and 18Z cycles are limited to shorter forecast ranges (up to $+90$ h). To maintain a fair comparison across the full 10-day forecast horizon, only the 00Z and 12Z initialization cycles were retained for both models, with forecasts evaluated every 6 hours from $t=+6$ h to $t=+240$ h.

The IFS HRES forecasts and the operational IFS analysis ($t=0$ frame) were retrieved via the ECMWF Open Data API~\cite{ECMWF-opendata}. To ensure complete data availability for our selected temporal domains, the extraction was routed specifically through the Amazon Web Services (AWS) mirror, which provided a more complete repository for the selected months than the native ECMWF, GCS, or Azure nodes.
    
The GraphCast (operational) forecasts were sourced directly from Google DeepMind's WeatherNext 1 public Google Cloud Storage buckets~\cite{google2025weathernext}.

\subsubsection{Data Harmonization and ZARR Consolidation} 
Because the source datasets were already provided at the correct spatial and temporal resolutions, the data engineering effort focused on structural harmonization. The data ingestion and transformation were executed using Google Colab Pro environments. Three distinct Python extraction modules were developed to handle the specific idiosyncrasies of the three target datasets: the IFS analysis, the IFS HRES forecasts, and the GraphCast operational forecasts. The processing pipeline performed the following critical operations on a day-by-day basis:

    - Variable Selection and Renaming: Parsing the downloaded GRIB2 files to isolate the specific variables ($T_{850}$, $Q_{850}$, and $Z_{500}$) and renaming them to establish a uniform nomenclature across all models.
    
    - Unit Conversion: The IFS HRES native output provides Geopotential Height. To match the strict input requirements of the Weatherbench evaluation framework, a post-processing step was implemented to convert Geopotential Height into Geopotential by multiplying the arrays by the standard gravity constant recommended by ECMWF ($g = 9.80665~m/s^2$)~\cite{ecmwf_gh}.
    
    - Dimensional Alignment: Standardizing coordinate systems and dimension names to ensure perfect tensor alignment between the NWP and ML outputs.
    
    - Metadata Consolidation: Once harmonized, the data was serialized from GRIB2 into ZARR, a format highly optimized for chunked, cloud-native multidimensional arrays. Crucially, this step included the consolidation of all metadata files. This consolidated metadata structure drastically reduces the overhead of cloud storage requests, allowing for significantly faster read-access during the evaluation phase.

This pipeline successfully generated 27 distinct, fully harmonized ZARR datasets, encompassing approximately 145 GB of data, which were subsequently hosted in a dedicated personal GCS bucket.

\subsubsection{Evaluation Execution} 
Once the forecast and reference datasets were consolidated into cloud-hosted ZARR stores, the evaluation workflow was executed in two stages.

First, the forecasts were evaluated using the WeatherBench-X framework against the corresponding ECMWF analysis data, which served as the reference representation of the atmospheric state. Deterministic verification metrics were computed for each forecast lead time and subsequently aggregated over the predefined regional domains, providing a consistent assessment of model performance across the different atmospheric regimes considered in this study.

The resulting metric datasets were then analyzed to characterize the comparative predictive skill of the forecasting systems. Regional and lead-time-dependent statistics were summarized through a series of visualizations designed to highlight differences in forecast accuracy, error growth, and overall model behavior across the evaluated variables and geographical domains.

\subsection{Evaluation Framework - Weatherbench-X} 
\vspace{1em}
The quantitative evaluation in this study is executed utilizing WeatherBench-X, the latest iteration of the open-source evaluation framework, widely used to compare forecasting skills between NWP and MLWP models~\cite{Weatherbench-2}~\cite{Weatherbench-X}.

The WeatherBench initiative was specifically designed to accelerate progress in data-driven weather modeling by providing a common, highly reliable foundation for comparing emerging data-driven models against traditional numerical baselines. As the field of MLWP has evolved, this framework has become widely accepted by both the global meteorological and machine learning communities as the standard for conducting rigorous, unbiased comparative analyses. The robustness of the tool stems from its strict adherence to the established forecast verification practices and mathematical guidelines employed by ECMWF and the World Meteorological Organization (WMO)~\cite{Weatherbench-2}.

By leveraging WeatherBench-X, this methodology ensures that all evaluations are performed consistently across the 27 generated ZARR datasets. The framework provides a highly optimized, cloud-native environment engineered to efficiently process massive multidimensional tensors. This standardization guarantees that spatial complexities such as the geometric area weighting of spherical grid cells are handled natively and identically for all models, allowing for an authentic evaluation of the operational skill between the selected models.

\subsection{Evaluation Metrics}
\vspace{1em}
\subsubsection{Root mean squared error (RMSE)} 
The RMSE is a standard deterministic metric utilized to quantify the absolute magnitude of forecast errors. When evaluating atmospheric models on a global or large regional scale, the data is typically projected onto an equiangular latitude-longitude grid. Because the Earth is spherical, grid cells at the poles have a significantly smaller physical area compared to grid cells at the equator. Weighing all cells equally during the error aggregation would result in an inordinate spatial bias towards higher latitudes. To prevent this distortion, the RMSE calculation strictly incorporates an area-weighting mechanism, applying a latitude-based weight $w(i)$ to ensure that the error at each grid point is proportional to its actual physical area. Note that this agrees with the definition of the RMSE used by the WMO and ECMWF~\cite{Lam2023}~\cite{Weatherbench-2}.

\begin{table}[ht]
\centering
\caption{Notation used in the evaluation metrics~\cite{Weatherbench-2}.}
\label{tab:notation}

\begin{tabular}{@{}lll@{}}
\toprule
\textbf{Symbol} & \textbf{Range} & \textbf{Description} \\
\midrule
$f$             &                & Forecast \\
$o$             &                & Ground Truth \\
$c$             &                & Climatology \\
$t$             & $1,\dots,T$    & Verification time \\
$l$             & $1,\dots,L$    & Lead time \\
$i$             & $1,\dots,I$    & Latitude index \\
$j$             & $1,\dots,J$    & Longitude index \\
$m$             & $1,\dots,M$    & Ensemble member index \\
\bottomrule
\end{tabular}
\end{table}

\begin{equation}\label{eq:eq1}
RMSE_{l} = \sqrt{ \frac{1}{T I J} \sum_{t=1}^{T} \sum_{i=1}^{I} \sum_{j=1}^{J} w(i) (f_{t,l,i,j} - o_{t,i,j})^{2} }
\end{equation}
\vspace{1em}

\subsubsection{Anomaly Correlation Coefficient (ACC)} 
While the RMSE evaluates the absolute magnitude of forecast errors, it does not fully capture a model's ability to predict the spatial structure of weather systems. To evaluate the spatial phase accuracy of the forecasts, this study utilizes the Anomaly Correlation Coefficient. The ACC is computed as the latitude-weighted Pearson correlation coefficient of the forecast and observation anomalies with respect to a defined climatological baseline.

This metric is highly valued in operational meteorology because it quantifies how well a model predicts the pattern and positioning of synoptic-scale features—such as advancing troughs, ridges, and pressure systems—independent of systematic biases in their absolute amplitude. The ACC score ranges from 1 (indicating perfect spatial correlation) to -1 (indicating perfect anti-correlation). A forecast that merely predicts the climatological average will yield an ACC value of 0. Operationally, the ECMWF considers an ACC value of 0.60 as the critical threshold~\cite{Weatherbench-2}. Below this value, the positioning of synoptic-scale features ceases to have practical value for forecasting purposes.

To compute the ACC, the atmospheric anomalies for both the forecast ($f^{\prime}$) and the ground truth observation ($o^{\prime}$) must first be defined by subtracting the climatological mean ($c$)~\cite{Lam2023}~\cite{Weatherbench-2}:

\begin{equation}\label{eq:eq2}
f_{t,l,i,j}^{\prime} = f_{t,l,i,j} - c_{t,l,i,j}
\end{equation}
\begin{equation}\label{eq:eq3}
o_{t,i,j}^{\prime} = o_{t,i,j} - c_{t,i,j}
\end{equation}

\vspace{1em}
The latitude-weighted ACC for a specific lead time across the evaluation domain is then formalized as~\cite{Lam2023}~\cite{Weatherbench-2}:

\begin{equation}\label{eq:eq4}
ACC_{l} = \frac{1}{T} \sum_{t=1}^{T} \frac{ \sum_{i=1}^{I} \sum_{j=1}^{J} w(i) f_{t,l,i,j}^{\prime} o_{t,i,j}^{\prime} }{ \sqrt{ \sum_{i=1}^{I} \sum_{j=1}^{J} w(i) (f_{t,l,i,j}^{\prime})^{2} \sum_{i=1}^{I} \sum_{j=1}^{J} w(i) (o_{t,i,j}^{\prime})^{2} } }
\end{equation}

\subsubsection{Climatology Dataset} 
To accurately calculate the ACC and establish a baseline for spatial anomalies, this study utilizes the standardized climatology dataset inherently provided by the WeatherBench-X framework. This baseline dataset is derived from a 30-year historical average of the ERA5 reanalysis, spanning the continuous period from 1990 to 2019. Relying on this pre-computed, long-term climatology ensures that the calculated spatial anomalies are consistent with established benchmarking protocols and effectively isolate true weather signals from short-term atmospheric noise~\cite{Weatherbench-2}.

\subsubsection{Normalized RMSE Difference} 
While RMSE and ACC provide essential absolute measures of model performance, directly comparing absolute errors across disparate meteorological variables—such as Geopotential Height evaluated in $m^2/s^2$ versus Specific Humidity evaluated in $g/kg$ can obscure the relative performance gains. To establish a unified, easily interpretable evaluation of GraphCast's operational skill relative to the IFS HRES baseline, this study calculates the Normalized RMSE Difference.

This comparative metric isolates the relative percentage gain or degradation in predictive skill of the artificial intelligence model against the traditional numerical benchmark. By normalizing the error difference against the baseline's own error, it standardizes the scale across all variables, regions, and lead times. The Normalized RMSE Difference (ND) is calculated using the following formula~\cite{Lam2023}:

\vspace{1em}
\begin{equation}\label{eq:eq5}
ND = \left( \frac{RMSE_{GraphCast} - RMSE_{IFS\_HRES}}{RMSE_{IFS\_HRES}} \right) \times 100
\end{equation}

\vspace{1em}
The mathematical orientation of this metric is highly intuitive for comparative analysis:

    - Negative Values: A negative normalized difference indicates a reduction in error relative to the baseline. This means that GraphCast outperforms the IFS HRES, exhibiting superior predictive skill.
    
    - Positive Values: A positive normalized difference signifies an increase in error, indicating that the traditional numerical method retains superior predictive skill for that specific variable and lead time.
    
    - Zero: A value of zero indicates parity in forecasting skill between the two models.

Inverted Representation (Baseline = 100\%): it is important to note that in certain visual representations and benchmarking scorecards, this relative performance is sometimes converted into an inverted Skill Score where the reference model's skill (e.g., IFS HRES) is set exactly at 100\%. In this specific formatting, an evaluated skill value greater than 100\% signifies that the AI model possesses better predictive skill (a lower relative error) than the baseline, whereas a value below 100\% represents worse predictive skill~\cite{Hoyer-SotA}~\cite{RASP-SotA}.

\begin{equation}\label{eq:eq6}
Skill\_Score (\%) = 100\% - ND
\end{equation}

\vspace{1em}
By expressing these comparative differences as a percentage, we can effectively aggregate, compare, and visualize the relative strengths and weaknesses of the MLWP paradigm across the distinct climatic zones of the Brazilian spatial domain.

%% file: sections/results.tex
\section{Results}
\vspace{1em}

This section evaluates the forecast skill of GraphCast operational against the ECMWF IFS HRES baseline over Brazil. Following the protocol established in the preview section, both models are verified against the same ground truth (the analysis data used to initialize both models at t=0) using the WeatherBench-X framework. We calculate RMSE and ACC across four climatic subregions (North, Northeast, Center-West/Southeast, and South), four seasonal months (March, July, October 2024; January 2025), and three variables: geopotential at 500 hPa (Z500), temperature at 850 hPa (T850), and specific humidity at 850 hPa (Q850), at lead times from 6h to 240h.

\begin{figure}[H]
\centering

\subcaptionbox{Mar/24 - summer-autumn (transition)\label{fig:T850-SS-a}}{
    \includegraphics[width=0.7\linewidth]{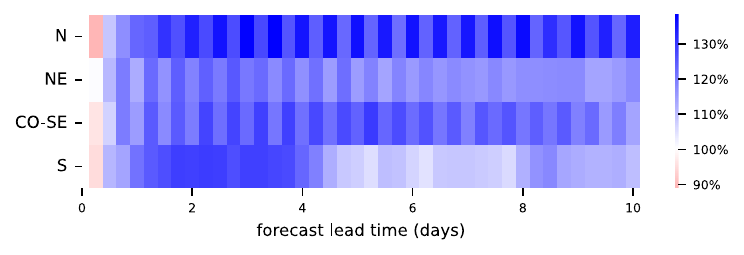}
}

\vspace{0.5em}
\subcaptionbox{Jul/24 - winter\label{fig:T850-SS-b}}{
    \includegraphics[width=0.7\linewidth]{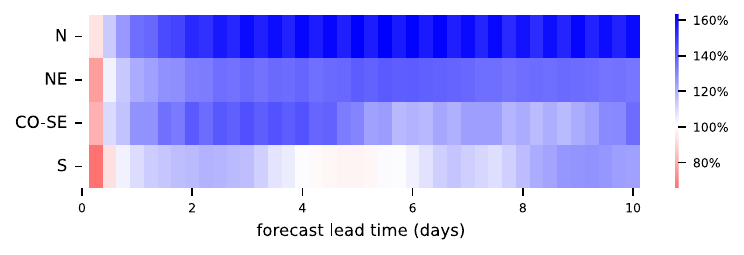}
}

\vspace{0.5em}
\subcaptionbox{Oct/24 - spring \label{fig:T850-SS-c}}{
    \includegraphics[width=0.7\linewidth]{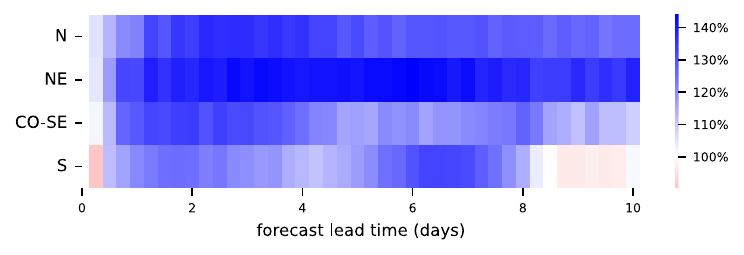}
}

\vspace{0.5em}
\subcaptionbox{Jan/25 - summer \label{fig:T850-SS-d}}{
    \includegraphics[width=0.7\linewidth]{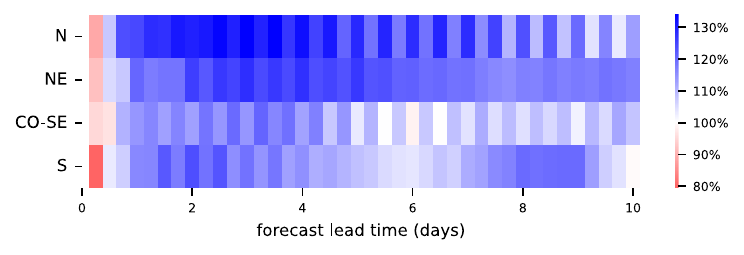}
}

\caption{$T_{850}$ skill score heatmaps, four seasonal months, and four subregions (N, NE, CO-SE, S). Source: authors.}
\label{fig:T850-SS}
\end{figure}

\begin{figure}[H]
\centering

\subcaptionbox{Mar/24 - summer-autumn (transition)\label{fig:Q850-SS-a}}{
    \includegraphics[width=0.7\linewidth]{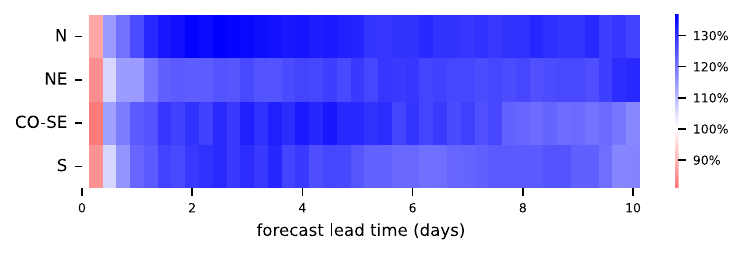}
}

\vspace{0.5em}
\subcaptionbox{Jul/24 - winter\label{fig:Q850-SS-b}}{
    \includegraphics[width=0.7\linewidth]{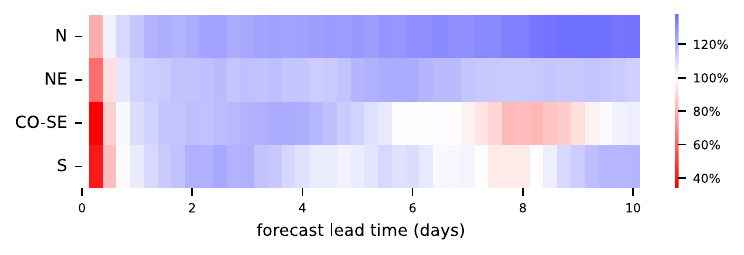}
}

\vspace{0.5em}
\subcaptionbox{Oct/24 - spring \label{fig:Q850-SS-c}}{
    \includegraphics[width=0.7\linewidth]{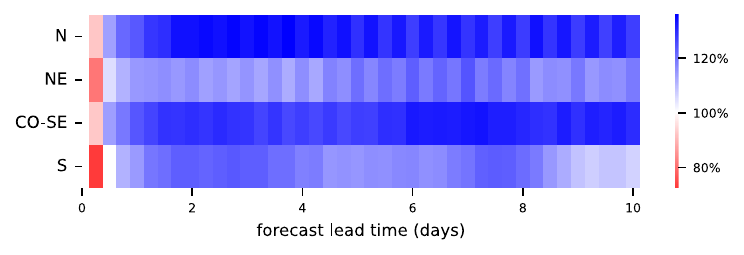}
}

\vspace{0.5em}
\subcaptionbox{Jan/25 - summer \label{fig:Q850-SS-d}}{
    \includegraphics[width=0.7\linewidth]{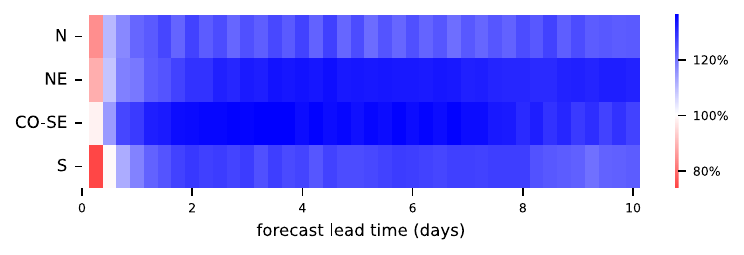}
}

\caption{$Q_{850}$ skill score heatmaps, four seasonal months, and four subregions (N, NE, CO-SE, S). Source: authors.}
\label{fig:Q850-SS}
\end{figure}

\begin{figure}[H]
\centering

\subcaptionbox{Mar/24 - summer-autumn (transition)\label{fig:Z500-SS-a}}{
    \includegraphics[width=0.7\linewidth]{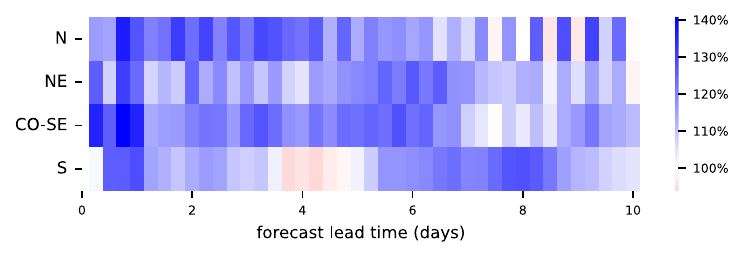}
}

\vspace{0.5em}
\subcaptionbox{Jul/24 - winter\label{fig:Z500-SS-b}}{
    \includegraphics[width=0.7\linewidth]{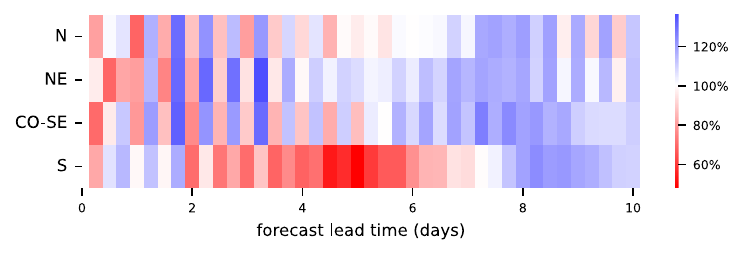}
}

\vspace{0.5em}
\subcaptionbox{Oct/24 - spring \label{fig:Z500-SS-c}}{
    \includegraphics[width=0.7\linewidth]{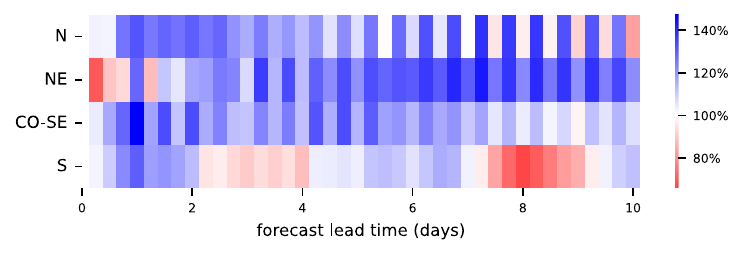}
}

\vspace{0.5em}
\subcaptionbox{Jan/25 - summer \label{fig:Z500-SS-d}}{
    \includegraphics[width=0.7\linewidth]{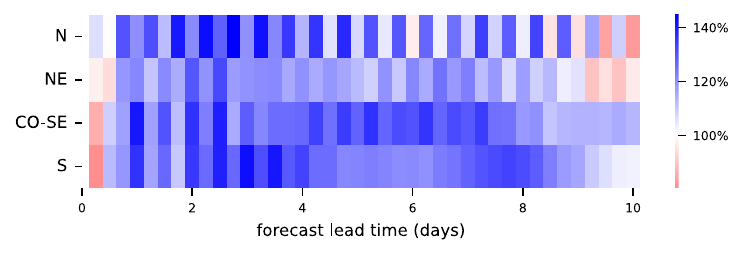}
}

\caption{$Z_{500}$ skill score heatmaps, four seasonal months, and four subregions (N, NE, CO-SE, S). Source: authors.}
\label{fig:Z500-SS}
\end{figure}

Figures~\ref{fig:T850-SS}, \ref{fig:Q850-SS} and \ref{fig:Z500-SS} present the Skill Score heatmaps for all three variables, covering the four months and four subregions (Y-axis). The overall pattern is immediately visible: in March, October, and January, GraphCast operational achieves Skill Scores consistently above the IFS baseline across all variables, regions, and lead times. July 2024 is the only month that deviates, and the deviation is specific to Z500 in the South, where the Skill Score drops to 47.91\% at 120h lead time (52.09\% below parity). This is the most severe deficit observed in any combination studied. For T850 and Q850, the Skill Score remains above the baseline even in July, indicating that the degradation is confined to the dynamic circulation field.

The most prominent departure from the baseline pattern thus occurs in the South during the austral winter, where Z500 shows systematic and severe degradation between lead times 48h and 168h (days 2 to 7). We examine this deficit and then contrast its seasonal and regional signatures with the behavior of the tropical subregions.

\subsection{Z500 in the South during July/24}

The $Z_{500}$ skill score distributions (Figure~\ref{fig:Z500-SS}) reveal that GraphCast generally maintains competitive performance relative to IFS HRES across most regions and seasons, consistent with the importance of 500 hPa flow in representing the large-scale baroclinic dynamics that govern subtropical and temperate weather patterns in South America~\cite{Coiffier2011-jz, Webster2020-xo}. However, the results also expose a distinct seasonal and regional degradation in GraphCast skill during July 2024, particularly over the South subregion, where extended areas of negative skill score anomalies (SS < 100\%) become evident.

This pronounced reduction in relative skill indicates that, under specific wintertime circulation regimes affecting southern Brazil, GraphCast exhibits larger forecast errors than the HRES baseline at medium and extended lead times. The spatial and temporal coherence of this signal suggests that the degradation is not associated with isolated forecast cases, but rather with systematic difficulties in representing the synoptic-scale dynamics characteristic of this period.

Table~\ref{tab:z500-south-ss} quantifies forecasting skills compared to the baseline across all four seasons. March, October, and January show SS consistently above 100\% (102\% to 135\%) at every lead time, indicating that GraphCast maintains lower RMSE than IFS HRES in the South during autumn, spring, and summer. The behavior in July is markedly different. At 12 h the SS is near parity (105.89\%), then drops to 70.01\% at 48 h and reaches 47.91\% at 120 h — the lowest SS observed in any variable-region-season combination in this study, corresponding to a deficit of 52.09\% below the IFS baseline. Recovery begins at 168 h (93.02\%), with SS crossing above 100\% at approximately 174 h (day 7.25). The model then reaches a post-recovery maximum of 123.29\% at 198 h (day 8.25) and maintains SS above 100\% through 240 h (108.95\%).

\begin{table}[ht]
\centering
\caption{Skill Score (\%) for Z500 in the South subregion (brazil-s) across the four seasonal months. SS $<$ 100 indicates GraphCast under performs the IFS HRES baseline.}
\label{tab:z500-south-ss}
\begin{tabular}{lccccc}
\toprule
\textbf{Lead time} & \textbf{Mar 2024} & \textbf{Jul 2024} & \textbf{Oct 2024} & \textbf{Jan 2025} \\
\midrule
12h          & 125.55 & 105.89 & 109.31 & 111.89 \\
48h  (day 2) & 113.44 & \textbf{70.01}  & 112.39 & 134.70 \\
120h (day 5) & 102.23 & \textbf{47.91}  & 103.16 & 121.33 \\
168h (day 7) & 119.55 & \textbf{93.02}  & 102.43 & 127.57 \\
240h (day 10)& 104.18 & 108.95 & 111.55 & 102.32 \\
\bottomrule
\end{tabular}
\end{table}

\begin{figure}[H]
  \centering
  \includegraphics[width=0.7\linewidth]{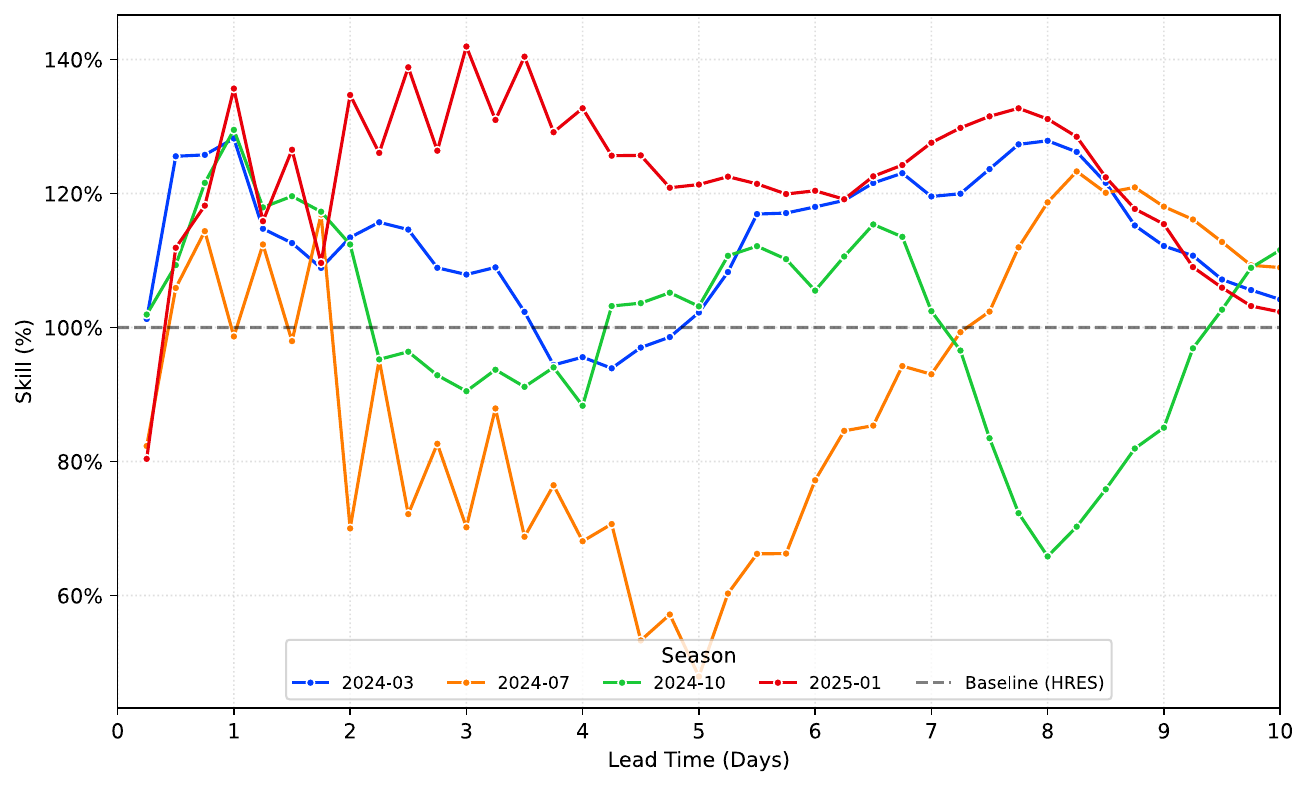}
\caption{Seasonal Skill Score for $Z_{500}$ in the South subregion. The orange (July) and green (October) lines denote the pronounced degradation of AI predictive skill during baroclinic seasons. Source: authors}
\label{fig:fig6}
\end{figure}

The seasonal SS curves in Figure~\ref{fig:fig6} confirm that July is the outlier: the remaining three months remain close to or above the 100\% parity line for most of the forecast range, while July diverges sharply between 48 and 168 h. The October curve shows a transient dip below 100\% near day 8, but examination of the underlying RMSE trajectories reveals that this feature does not reflect a genuine loss of GraphCast skill. Instead, it coincides with an anomalous decrease in IFS HRES RMSE at those lead times (a non-monotonic behavior inconsistent with expected NWP error growth). Because the focus of this study is on systematic patterns in GraphCast performance, we do not pursue this artifact further here.

\subsubsection{Differential Skill Degradation Across Atmospheric Variables During July 2024
}

\begin{figure}[H]
  \centering
  \includegraphics[width=0.7\linewidth]{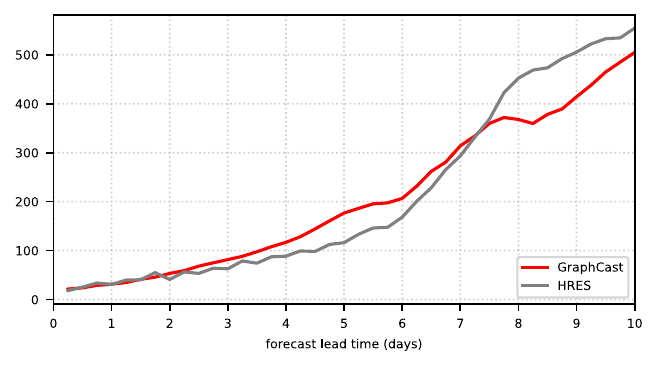}\\
  \includegraphics[width=0.7\linewidth]{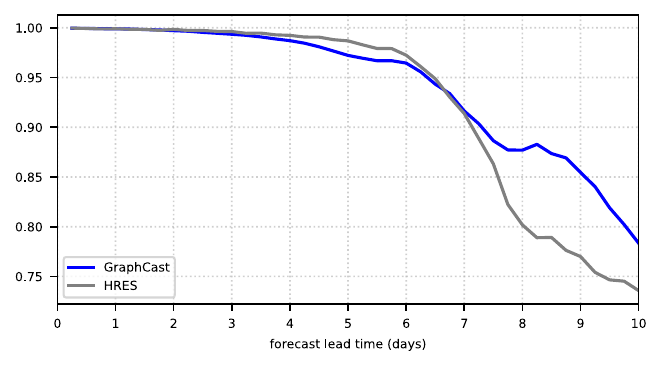}
\caption{Comparative performance metrics for $Z_{500}$ in the South subregion (July 2024). Top: RMSE demonstrating the medium-range divergence. Bottom: ACC highlighting phase tracking capabilities. Source: authors}
\label{fig:stacked_rmse_acc}
\end{figure}

To diagnose the mechanism of the winter deficit, the Skill Scores for the three variables in the South during July 2024 are compared in Table~\ref{tab:vertical-selectivity} and in a vertically stacked layout of RMSE and ACC trajectories (Figure~\ref{fig:stacked_rmse_acc}).

Table~\ref{tab:vertical-selectivity} reveals a pronounced contrast: Z500 reaches a minimum of 47.91\% at 120 h, representing a deficit of 52.09\% below the IFS baseline. At the same lead time, however, T850 (97.28\%) and Q850 (105.00\%) remain near or above parity. The RMSE curves in Figure~\ref{fig:stacked_rmse_acc} confirm this divergence. The Z500 trajectories separate sharply between 48 and 168 h, with GraphCast exceeding 300 m²/s² while IFS HRES remains below 200 m²/s². Meanwhile, the 850 hPa variables maintain comparable error levels between the two models.

\begin{table}[ht]
\centering
\caption{Skill Score (\%) for the three variables in the South subregion during July 2024.}
\label{tab:vertical-selectivity}
\begin{tabular}{lcccccc}
\toprule
\textbf{Variable} & \textbf{Level} & \textbf{12h} & \textbf{48h} & \textbf{120h} & \textbf{168h} & \textbf{240h} \\
\midrule
\textbf{Z500} & 500 hPa & 105.89 & 70.01  & \textbf{47.91} & 93.02 & 108.95 \\
\textbf{T850} & 850 hPa & 92.29  & 117.24 & \textbf{97.28}  & 112.06 & 123.70 \\
\textbf{Q850} & 850 hPa & 83.22  & 120.31 & \textbf{105.00} & 102.78 & 119.31 \\
\bottomrule
\end{tabular}
\end{table}

The results indicate a clear disparity between the behavior of $Z_{500}$ and the thermodynamic variables evaluated in this study. While substantial skill degradation is observed for $Z_{500}$ during July 2024, comparable reductions are not evident in either $T_{850}$ or $Q_{850}$. This pattern suggests that the identified performance deficit is variable-dependent rather than uniformly distributed across all evaluated forecast fields.

\subsubsection{Regional contrasts in the austral winter}

The winter signature is not exclusive to the South. Table~\ref{tab:regional-winter} presents the Z500 Skill Scores for all four subregions during July 2024, revealing a two-part pattern. At 48 h, every region is below the 100\% parity line: the North (87.82\%), the Northeast (81.95\%), the Center-West/Southeast (76.24\%), and the South (70.01\%) — indicating that the austral winter imposes a short-range challenge on GraphCast across the whole of Brazil, regardless of climate regime.

The divergence is in the recovery. The tropical subregions (North and Northeast) return above parity by 120 h (96.14\% and 106.88\%, respectively), and the Center-West/Southeast follows by 168 h (111.70\%). The South is the exception: rather than recovering, the SS deepens from 70.01\% at 48 h to a minimum of 47.91\% at 120 h — the lowest value in the study — and does not return above parity until after 240 h.

This asymmetry has a structural basis in GraphCast's architecture. Bonavita~\cite{Bonavita2024} expresses that in autoregressive ML models like GraphCast, the divergent component of the flow is progressively suppressed relative to the rotational component, with spectral energy loss concentrating at sub-synoptic scales (~500–700 km) and compounding with each 6 hour rollout step. In tropical regimes, where dominant dynamics evolve at planetary and synoptic scales over days to weeks, this suppression is relatively benign. In the extratropical South in winter, where baroclinic wave development is sustained precisely by sub-synoptic divergent motions at the scales most affected, the 6 h timestep inherited from ERA5 archiving frequency structurally undersamples the dynamics it is asked to predict — and the deficit deepens rather than recovers.

\begin{table}[ht]
\centering
\caption{Skill Score (\%) for Z500 across the four subregions during July 2024.}
\label{tab:regional-winter}
\begin{tabular}{lccccc}
\toprule
\textbf{Region} & \textbf{12h} & \textbf{48h} & \textbf{120h} & \textbf{168h} & \textbf{240h} \\
\midrule
North           & 101.29 &  87.82 &  96.14 & 101.76 & 111.61 \\
Northeast       &  69.06 &  81.95 & 106.88 & 114.97 & 112.06 \\
Center-West/SE  &  95.70 &  76.24 &  86.75 & 111.70 & 110.23 \\
\textbf{South}  & \textbf{105.89} & \textbf{70.01} & \textbf{47.91} & \textbf{93.02} & \textbf{108.95} \\
\bottomrule
\end{tabular}
\end{table}

\subsubsection{The South in other seasons}

To complete the seasonal picture, Table~\ref{tab:south-120h-seasonal} summarizes the Z500 Skill Score at 120 h (the lead time of maximum degradation) across the four months. March (102.23\%) and October (103.16\%) are marginally above parity, indicating practical equivalence between the two models during autumn and spring. January reaches 121.33\%, the highest value of the four, confirming that GraphCast clearly outperforms IFS HRES during the summer in the extratropical South as well. The July minimum (47.91\%) stands apart: a deficit of 52.09\% below the baseline, more than an order of magnitude larger than any other seasonal departure from parity.

The seasonality of the deficit, maximum in winter and absent in the other three seasons is consistent with the annual cycle of baroclinic activity in mid-latitudes, which peaks during the winter of each hemisphere~\cite{Webster2020-xo}. This synchrony further supports the hypothesis that the fixed 6 h timestep of GraphCast, adequate for slower-evolving tropical and summer dynamics, is a structural limitation under winter baroclinic conditions. In contrast to the Z500 deficit, T850 and Q850 Skill Scores in the tropical subregions during July remain above the baseline at most lead times, consistent with the hypothesis that the 6 h timestep is adequate for the slower-evolving dynamics of the tropical lower troposphere.

\begin{table}[ht]
\centering
\caption{Skill Score (\%) for Z500 in the South subregion at 120 h lead time, by seasonal month.}
\label{tab:south-120h-seasonal}
\begin{tabular}{lcc}
\toprule
\textbf{Month} & \textbf{Season} & \textbf{SS 120h} \\
\midrule
Mar 2024 & Autumn  & 102.23 \\
\textbf{Jul 2024} & \textbf{Winter}  & \textbf{47.91} \\
Oct 2024 & Spring  & 103.16 \\
Jan 2025 & Summer  & 121.33 \\
\bottomrule
\end{tabular}
\end{table}

\subsubsection{Note to statistical significance}

To assess the robustness of the results, a two-sided paired t-test with AR(2) autocorrelation correction was applied following Geer~(2016)~\cite{Geer2016} and the GraphCast evaluation protocol~\cite{Lam2023}. The four seasonal months were treated as independent blocks: within each month, the 61 paired RMSE differences (GraphCast minus IFS HRES) form an autocorrelated time series, while no correlation is assumed between months. This block structure was used to estimate a pooled AR(2) autocorrelation structure, from which variance inflation factors were computed for each lead time ($k = 1.13$ to $2.72$, corresponding to effective sample sizes $N_{\text{eff}} = 90$ to $215$ across the full 244-init dataset).

Across all four months pooled, GraphCast shows significantly lower RMSE than IFS HRES in the South at 12 h ($p < 0.001$), 48 h ($p < 0.05$), and 168 h ($p < 0.05$). At 120 h and 240 h, the mean difference is not statistically significant, indicating that the overall performance advantage at these lead times is not distinguishable from zero given the observed variability.

For July specifically, the month where the aggregate Skill Score reaches 47.91\% at 120 h — the paired difference is positive (GraphCast RMSE higher than IFS RMSE by 26.0 m$^2$/s$^2$ on average) but does not reach significance at the 0.05 level ($p = 0.16$). This reflects the high init-to-init variability of the RMSE at this lead time (standard deviation of 110.4 m$^2$/s$^2$) relative to the mean difference, which limits the power of the per-init test. The Skill Score, which pools over all 248 $\times$ 61 grid points simultaneously, provides a more powerful diagnostic and remains the primary metric for the analysis. Detailed results for selected lead times in Table~\ref{tab:significance2} and for all lead times for all three variables in the Appendix~\ref{subsec:significance-tables}.


\begin{table}[ht]
\centering
\caption{Pooled paired t-test with AR(2) correction for Z500 in the South subregion. The four seasonal months (244 init times) are treated as independent blocks with a pooled AR(2) autocorrelation structure. $k$ is the variance inflation factor; $N_{\text{eff}}$ is the total effective sample size. Significance levels: $p < 0.05$ (*), $p < 0.01$ (**), $p < 0.001$ (***); ns = not significant.}
\label{tab:significance2}
\begin{tabular}{lcccccc}
\toprule
\textbf{Lead time} & \textbf{$k$} & \textbf{$N_{\text{eff}}$} & \textbf{$\bar{d}$} (m$^2$/s$^2$) & \textbf{$t$} & \textbf{$p$} & \textbf{Sig} \\
\midrule
12h   & 1.18 & 206 &  $-$3.7  & $-$3.58 & 0.0004 & *** \\
48h   & 2.51 & 97  &  $-$8.7  & $-$2.50 & 0.0142 & *   \\
120h  & 1.64 & 149 &  $-$3.8  & $-$0.57 & 0.5695 & ns  \\
168h  & 1.57 & 156 & $-$24.3  & $-$2.18 & 0.0306 & *   \\
240h  & 1.13 & 215 & $-$25.7  & $-$1.50 & 0.1345 & ns  \\
\bottomrule
\end{tabular}
\end{table}

%% file: sections/conclusions.tex
\section{CONCLUSIONS}

The observed pattern, consistent skill in the tropics and systematic degradation in the extratropics during winter, aligns with the fundamental differences in atmospheric dynamics between these regimes. Tropical circulation is dominated by deep convection and equatorial waves, phenomena that evolve over days to weeks. In mid-latitudes, dynamics are governed by baroclinic instability, with systems developing and propagating on timescales of 12 to 48 h.

The 6 h temporal resolution of GraphCast operational, inherited from the ERA5 archiving frequency, is sufficient for the slow evolution of tropical phenomena but appears inadequate to resolve the growth and propagation of baroclinic systems. This finding is consistent with the original GraphCast evaluation~\cite{Lam2023}, which reported slightly lower performance in the Southern Hemisphere relative to the Northern Hemisphere, and with recent studies identifying temporal resolution as a key bottleneck for MLWP models in mid-latitudes~\cite{ Chantry2024}.

The vertical decoupling documented in this work, severe error concentrated at 500 hPa and virtually absent at 850 hPa, imposes a strong constraint on possible explanations. The root cause must be specific to the dynamical regime at upper levels like Rossby waves and baroclinic instability, and does not affect the lower-tropospheric dynamics, which are topographically blocked and thermodynamically forced~\cite{Webster2020-xo}. This excludes explanations based on generic model deficiencies that would affect all levels equally, such as insufficient horizontal resolution or lack of physical parameterizations, and points to causes related to the temporal sampling of baroclinic evolution.

\subsection{Hypotheses for the South bias}

Two complementary hypotheses, one dynamical and the other orographic, might explain the observed pattern:

\textbf{$H_{1}$} - Insufficient timestep for baroclinic evolution. The fixed 6 h timestep of GraphCast undersamples the growth rates of the most unstable baroclinic modes, whose e-folding time in the austral winter is 12 to 24 h. With only 2 to 4 samples per trough life cycle, the model accumulates phase error that manifests as severe degradation in the 500 hPa geopotential, the level where synoptic-scale circulation is free and unconstrained by surface processes.

\textbf{$H_{2}$} - Topographic blocking by the Andes as a vertical decoupling barrier. The Andes, with crest heights of approximately 3,000 to 4,000 m in the South Brazil and northern Argentina region, physically separate the dynamics at 500 hPa (above the crest, free flow) from those at 850 hPa (below the crest, channeled flow). This orographic barrier explains why the phase error at 500 hPa does not project onto T850 and Q850, fields that evolve under distinct topographic and thermodynamic constraints. Testing this hypothesis would require vertical profiles of RMSE at sufficient resolution to identify the error discontinuity at the Andean crest height.

Additional factors that may contribute, though likely secondary, include the possible degradation of ERA5 analysis quality in the Southern Hemisphere (lower in situ observation density) and the difference in native horizontal resolution between GraphCast (0.25°) and IFS HRES (~9 km), which, although both compared at 0.25° grid spacing, may still favor IFS HRES in the representation of frontal gradients. The detailed investigation of these mechanisms is beyond the scope of this work.

\subsection{Limitations of the proposed scope}

Three main limitations bound the generalizability of these results:

- Temporal coverage: Four seasonal months (124 days) were analyzed, covering all four seasons but not interannual variability. Years modulated by ENSO may present distinct bias patterns, particularly in the South.

- Variable and model scope: The analysis was limited to three variables (Z500, T850, Q850) and the deterministic GraphCast operational model. Precipitation, wind, and surface variables were not included, nor were probabilistic models (GenCast) or other MLWP architectures (Pangu-Weather, FourCastNet).

- Fairness of comparison: The ground truth is the IFS HRES analysis itself, creating an intrinsic advantage for the numerical model, which is verified against its own analysis, over GraphCast, which is verified against an analysis external to its training. The impact of this asymmetry on the magnitude of Skill Scores cannot be quantified with the available data.

\subsection{Synthesis}

The results show that GraphCast performs comparably to or better than IFS HRES in the tropical Brazilian subregions and in the South during summer, autumn, and spring, with Skill Scores consistently above 100\% (102\% to 135\%) at all lead times. The sole exception is the South during the austral winter (July 2024), where GraphCast shows severe degradation in Z500: SS = 47.91\% at 120 h (52.09\% below parity). The deficit persists from 48 h to 174 h, with a crossing point of approximately day 7.5 — roughly 3 days later than in the tropical subregions. After recovery, GraphCast reaches a maximum of 123.29\% at 198 h (day 8.25) and maintains SS above 100\% through the end of the forecast cycle. The deficit is seasonal (maximum in winter, absent in other seasons), regional (exclusive to the South — the remaining subregions recover above 100\% by 120 h), variable-specific (severe in Z500, marginal in T850, robust in Q850), and dynamical (phase error in baroclinic systems, not thermodynamic error).

The key diagnostic signature of vertical selectivity, error concentrated at 500 hPa and absent at 850 hPa, distinguishes phase error in upper-level baroclinic dynamics from generic thermodynamic error, and points to the fixed 6 h timestep as the most probable root cause.

\subsection{Future Work}

Expanding the analysis to multiple years would allow the assessment of interannual variability (ENSO) on the documented bias. The inclusion of additional variables, such as precipitation, wind at surface and upper levels, shortwave radiation would broaden the relevance of the benchmark for sectoral applications. The sawtooth verification artifact mentioned would benefit from hourly temporal resolution to decouple phase error from amplitude error at short lead times. Finally, the localized and seasonal nature of the South deficit makes this subregion a natural target for fine-tuning or regional adaptation of GraphCast to extratropical winter conditions.

%% file: sections/appendix.tex
\newpage
\onecolumn

\appendix
\section{Appendix}

\subsection{Statistical significance tables}
\label{subsec:significance-tables}

\AtBeginEnvironment{longtable}{%
  \small
  \setlength{\tabcolsep}{5pt}%
  \renewcommand{\arraystretch}{0.95}%
}

\begin{longtable}{lcccccccc}
\caption{Pooled paired t-test with AR(2) correction for $Z_{500}$ in the South subregion of Brazil across all lead times. Negative $\bar{d}$ = GraphCast is better. Lead (h/d) = forecast lead time in hours and days; $k$ = AR(2) variance inflation factor; $N_{\text{eff}}$ = effective sample size across all 4 months; $\bar{d}$ = mean paired RMSE difference (GC $-$ IFS), in m$^2$/s$^2$; $\sigma_d$ = standard deviation of paired differences; $t$ = paired $t$-test statistic; $p$ = $p$-value; Sig = significance ($p < 0.05$ (*), $p < 0.01$ (**), $p < 0.001$ (***), ns = not significant).} \label{tab:z500_south}\\
\toprule
Lead (h) & Lead (d) & $k$ & $N_{\text{eff}}$ & $\bar{d}$ (m$^2$/s$^2$) & $\sigma_d$ (m$^2$/s$^2$) & $t$ & $p$ & Sig \\
\midrule
\endfirsthead
\toprule
Lead (h) & Lead (d) & $k$ & $N_{\text{eff}}$ & $\bar{d}$ (m$^2$/s$^2$) & $\sigma_d$ (m$^2$/s$^2$) & $t$ & $p$ & Sig \\
\midrule
\endhead
\midrule                         
\endfoot
\bottomrule                      
\endlastfoot
  6 & 0.25 & 1.46 & 167 & $+$1.7  &   11.2 & $+$1.92 & 0.0565 & ns  \\
 12 & 0.50 & 1.18 & 206 & $-$3.7  &   14.8 & $-$3.58 & $<0.001$ & *** \\
 18 & 0.75 & 1.80 & 135 & $-$7.7  &   19.0 & $-$4.72 & $<0.001$ & *** \\
 24 & 1.00 & 1.80 & 136 & $-$10.2 &   22.5 & $-$5.28 & $<0.001$ & *** \\
 30 & 1.25 & 2.59 &  94 & $-$6.2  &   22.5 & $-$2.66 & 0.0094 & **  \\
 36 & 1.50 & 2.00 & 122 & $-$6.9  &   26.1 & $-$2.90 & 0.0045 & **  \\
 42 & 1.75 & 2.68 &  91 & $-$8.3  &   29.6 & $-$2.68 & 0.0088 & **  \\
 48 & 2.00 & 2.51 &  97 & $-$8.7  &   34.3 & $-$2.50 & 0.0142 & *   \\
 54 & 2.25 & 2.68 &  91 & $-$8.5  &   36.4 & $-$2.22 & 0.0292 & *   \\
 60 & 2.50 & 2.59 &  94 & $-$9.7  &   40.2 & $-$2.34 & 0.0213 & *   \\
 66 & 2.75 & 2.27 & 108 & $-$6.3  &   40.8 & $-$1.60 & 0.1134 & ns  \\
 72 & 3.00 & 2.51 &  97 & $-$8.1  &   43.7 & $-$1.83 & 0.0706 & ns  \\
 78 & 3.25 & 2.24 & 109 & $-$8.9  &   44.1 & $-$2.10 & 0.0379 & *   \\
 84 & 3.50 & 2.51 &  97 & $-$7.3  &   47.9 & $-$1.51 & 0.1351 & ns  \\
 90 & 3.75 & 2.42 & 101 & $-$4.8  &   51.5 & $-$0.93 & 0.3531 & ns  \\
 96 & 4.00 & 2.36 & 103 & $-$4.6  &   55.6 & $-$0.84 & 0.4057 & ns  \\
102 & 4.25 & 2.35 & 104 & $-$6.2  &   58.0 & $-$1.08 & 0.2829 & ns  \\
108 & 4.50 & 1.68 & 145 & $-$3.4  &   68.9 & $-$0.60 & 0.5527 & ns  \\
114 & 4.75 & 1.70 & 143 & $-$7.0  &   73.2 & $-$1.14 & 0.2566 & ns  \\
\textbf{120} & \textbf{5.00} & \textbf{1.64} & \textbf{149} & \textbf{$-$3.8}  &   \textbf{80.8} & \textbf{$-$0.57} & \textbf{0.5695} & \textbf{ns}  \\
126 & 5.25 & 1.41 & 173 & $-$10.9 &   86.1 & $-$1.67 & 0.0968 & ns  \\
132 & 5.50 & 1.99 & 123 & $-$12.8 &   86.3 & $-$1.65 & 0.1022 & ns  \\
138 & 5.75 & 1.71 & 143 & $-$12.0 &   93.2 & $-$1.54 & 0.1258 & ns  \\
144 & 6.00 & 1.61 & 151 & $-$13.2 &  101.3 & $-$1.61 & 0.1102 & ns  \\
150 & 6.25 & 1.43 & 171 & $-$18.6 &  111.4 & $-$2.19 & 0.0302 & *   \\
156 & 6.50 & 1.39 & 176 & $-$20.5 &  123.3 & $-$2.20 & 0.0292 & *   \\
162 & 6.75 & 1.42 & 172 & $-$25.9 &  129.7 & $-$2.61 & 0.0099 & **  \\
168 & 7.00 & 1.57 & 156 & $-$24.3 &  139.0 & $-$2.18 & 0.0306 & *   \\
174 & 7.25 & 1.87 & 131 & $-$27.3 &  145.2 & $-$2.15 & 0.0334 & *   \\
180 & 7.50 & 1.79 & 137 & $-$30.2 &  151.4 & $-$2.33 & 0.0212 & *   \\
186 & 7.75 & 1.67 & 146 & $-$35.8 &  161.7 & $-$2.68 & 0.0083 & **  \\
192 & 8.00 & 1.46 & 168 & $-$37.4 &  172.2 & $-$2.82 & 0.0055 & **  \\
198 & 8.25 & 1.56 & 156 & $-$37.9 &  186.6 & $-$2.54 & 0.0120 & *   \\
204 & 8.50 & 1.84 & 133 & $-$36.0 &  186.5 & $-$2.22 & 0.0280 & *   \\
210 & 8.75 & 2.13 & 115 & $-$35.7 &  188.7 & $-$2.03 & 0.0449 & *   \\
216 & 9.00 & 1.80 & 136 & $-$28.7 &  191.6 & $-$1.74 & 0.0838 & ns  \\
222 & 9.25 & 1.50 & 163 & $-$22.8 &  212.7 & $-$1.37 & 0.1721 & ns  \\
228 & 9.50 & 1.21 & 201 & $-$21.2 &  240.4 & $-$1.25 & 0.2130 & ns  \\
234 & 9.75 & 1.13 & 215 & $-$19.2 &  247.5 & $-$1.14 & 0.2551 & ns  \\
240 & 10.00 & 1.13 & 215 & $-$25.7 &  250.9 & $-$1.50 & 0.1345 & ns  \\
\end{longtable}

\newpage
\begin{longtable}{lcccccccc}
\caption{Pooled paired t-test with AR(2) correction for $T_{850}$ in the South subregion of Brazil across all lead times. Negative $\bar{d}$ = GraphCast is better. Lead (h/d) = forecast lead time in hours and days; $k$ = AR(2) variance inflation factor; $N_{\text{eff}}$ = effective sample size across all 4 months; $\bar{d}$ = mean paired RMSE difference (GC $-$ IFS), in K; $\sigma_d$ = standard deviation of paired differences; $t$ = paired $t$-test statistic; $p$ = $p$-value; Sig = significance ($p < 0.05$ (*), $p < 0.01$ (**), $p < 0.001$ (***), ns = not significant).} \label{tab:t850_south}\\
\toprule
Lead (h) & Lead (d) & $k$ & $N_{\text{eff}}$ & $\bar{d}$ (K) & $\sigma_d$ (K) & $t$ & $p$ & Sig \\
\midrule
\endfirsthead
\toprule
Lead (h) & Lead (d) & $k$ & $N_{\text{eff}}$ & $\bar{d}$ (K) & $\sigma_d$ (K) & $t$ & $p$ & Sig \\
\midrule
\endhead
\midrule                         
\endfoot
\bottomrule                      
\endlastfoot
  6 & 0.25 & 2.85 &  86 & $+$0.0803 & 0.10 & $+$7.58 & $<0.001$ & *** \\
 12 & 0.50 & 3.37 &  72 & $-$0.0063 & 0.16 & $-$0.35 & 0.7303 &  ns \\
 18 & 0.75 & 1.94 & 126 & $-$0.0602 & 0.17 & $-$3.95 & $<0.001$ & *** \\
 24 & 1.00 & 3.15 &  77 & $-$0.1156 & 0.20 & $-$5.14 & $<0.001$ & *** \\
 30 & 1.25 & 2.35 & 104 & $-$0.1589 & 0.23 & $-$7.06 & $<0.001$ & *** \\
 36 & 1.50 & 2.13 & 114 & $-$0.1940 & 0.25 & $-$8.43 & $<0.001$ & *** \\
 42 & 1.75 & 2.68 &  91 & $-$0.2164 & 0.27 & $-$7.70 & $<0.001$ & *** \\
 48 & 2.00 & 2.12 & 115 & $-$0.2328 & 0.29 & $-$8.64 & $<0.001$ & *** \\
 54 & 2.25 & 3.04 &  80 & $-$0.2357 & 0.29 & $-$7.27 & $<0.001$ & *** \\
 60 & 2.50 & 2.56 &  95 & $-$0.2571 & 0.31 & $-$7.98 & $<0.001$ & *** \\
 66 & 2.75 & 2.71 &  90 & $-$0.2396 & 0.33 & $-$6.87 & $<0.001$ & *** \\
 72 & 3.00 & 2.53 &  96 & $-$0.2627 & 0.36 & $-$7.19 & $<0.001$ & *** \\
 78 & 3.25 & 2.83 &  86 & $-$0.2532 & 0.36 & $-$6.54 & $<0.001$ & *** \\
 84 & 3.50 & 2.95 &  83 & $-$0.2559 & 0.40 & $-$5.83 & $<0.001$ & *** \\
 90 & 3.75 & 3.03 &  81 & $-$0.2399 & 0.44 & $-$4.93 & $<0.001$ & *** \\
 96 & 4.00 & 3.32 &  73 & $-$0.2277 & 0.48 & $-$4.03 & $<0.001$ & *** \\
102 & 4.25 & 3.25 &  75 & $-$0.2117 & 0.54 & $-$3.38 & 0.0012 &  ** \\
108 & 4.50 & 3.00 &  81 & $-$0.2130 & 0.62 & $-$3.09 & 0.0027 &  ** \\
114 & 4.75 & 2.64 &  92 & $-$0.2232 & 0.66 & $-$3.24 & 0.0017 &  ** \\
120 & 5.00 & 2.22 & 110 & $-$0.2348 & 0.69 & $-$3.59 & $<0.001$ & *** \\
126 & 5.25 & 2.15 & 114 & $-$0.2470 & 0.74 & $-$3.56 & $<0.001$ & *** \\
132 & 5.50 & 2.25 & 108 & $-$0.2716 & 0.70 & $-$4.06 & $<0.001$ & *** \\
138 & 5.75 & 2.60 &  94 & $-$0.2794 & 0.68 & $-$3.97 & $<0.001$ & *** \\
144 & 6.00 & 2.65 &  92 & $-$0.2738 & 0.75 & $-$3.51 & $<0.001$ & *** \\
150 & 6.25 & 2.12 & 115 & $-$0.3091 & 0.79 & $-$4.18 & $<0.001$ & *** \\
156 & 6.50 & 1.84 & 133 & $-$0.3474 & 0.84 & $-$4.78 & $<0.001$ & *** \\
162 & 6.75 & 1.49 & 164 & $-$0.3688 & 0.89 & $-$5.30 & $<0.001$ & *** \\
168 & 7.00 & 1.35 & 181 & $-$0.3652 & 0.94 & $-$5.25 & $<0.001$ & *** \\
174 & 7.25 & 1.40 & 174 & $-$0.3550 & 0.93 & $-$5.02 & $<0.001$ & *** \\
180 & 7.50 & 1.53 & 160 & $-$0.3441 & 0.95 & $-$4.55 & $<0.001$ & *** \\
186 & 7.75 & 1.98 & 123 & $-$0.3407 & 1.07 & $-$3.54 & $<0.001$ & *** \\
192 & 8.00 & 1.71 & 142 & $-$0.3953 & 1.11 & $-$4.24 & $<0.001$ & *** \\
198 & 8.25 & 1.94 & 126 & $-$0.3982 & 1.15 & $-$3.87 & $<0.001$ & *** \\
204 & 8.50 & 1.69 & 145 & $-$0.4271 & 1.25 & $-$4.12 & $<0.001$ & *** \\
210 & 8.75 & 1.63 & 149 & $-$0.4037 & 1.35 & $-$3.66 & $<0.001$ & *** \\
216 & 9.00 & 1.47 & 166 & $-$0.4309 & 1.47 & $-$3.77 & $<0.001$ & *** \\
222 & 9.25 & 1.34 & 182 & $-$0.4210 & 1.60 & $-$3.55 & $<0.001$ & *** \\
228 & 9.50 & 1.17 & 208 & $-$0.3915 & 1.68 & $-$3.36 & $<0.001$ & *** \\
234 & 9.75 & 1.27 & 193 & $-$0.3935 & 1.72 & $-$3.17 & 0.0018 &  ** \\
240 & 10.00 & 1.28 & 191 & $-$0.4134 & 1.79 & $-$3.19 & 0.0016 &  ** \\
\end{longtable}

\newpage
\begin{longtable}{lcccccccc}
\caption{Pooled paired t-test with AR(2) correction for $q_{850}$ in the South subregion of Brazil across all lead times. Negative $\bar{d}$ = GraphCast is better. Lead (h/d) = forecast lead time in hours and days; $k$ = AR(2) variance inflation factor; $N_{\text{eff}}$ = effective sample size across all 4 months; $\bar{d}$ = mean paired RMSE difference (GC $-$ IFS), in g/kg; $\sigma_d$ = standard deviation of paired differences; $t$ = paired $t$-test statistic; $p$ = $p$-value; Sig = significance ($p < 0.05$ (*), $p < 0.01$ (**), $p < 0.001$ (***), ns = not significant).} \label{tab:q850_south}\\
\toprule
Lead (h) & Lead (d) & $k$ & $N_{\text{eff}}$ & $\bar{d}$ (g/kg) & $\sigma_d$ (g/kg) & $t$ & $p$ & Sig \\
\midrule
\endfirsthead
\toprule
Lead (h) & Lead (d) & $k$ & $N_{\text{eff}}$ & $\bar{d}$ (g/kg) & $\sigma_d$ (g/kg) & $t$ & $p$ & Sig \\
\midrule
\endhead
\midrule                         
\endfoot
\bottomrule                      
\endlastfoot
  6 & 0.25 & 2.85 &  86 & $+$0.0002 & 0.0001 & $+$13.85 & $<0.001$ & *** \\
 12 & 0.50 & 2.90 &  84 & $+$0.0000 & 0.0002 & $+$1.10 & 0.2761 &  ns \\
 18 & 0.75 & 2.31 & 106 & $-$0.0001 & 0.0002 & $-$4.32 & $<0.001$ & *** \\
 24 & 1.00 & 3.05 &  80 & $-$0.0002 & 0.0003 & $-$5.83 & $<0.001$ & *** \\
 30 & 1.25 & 2.85 &  85 & $-$0.0003 & 0.0003 & $-$7.75 & $<0.001$ & *** \\
 36 & 1.50 & 3.67 &  66 & $-$0.0003 & 0.0003 & $-$7.88 & $<0.001$ & *** \\
 42 & 1.75 & 3.33 &  73 & $-$0.0004 & 0.0003 & $-$9.35 & $<0.001$ & *** \\
 48 & 2.00 & 3.48 &  70 & $-$0.0004 & 0.0004 & $-$9.08 & $<0.001$ & *** \\
 54 & 2.25 & 3.21 &  76 & $-$0.0004 & 0.0004 & $-$9.45 & $<0.001$ & *** \\
 60 & 2.50 & 2.82 &  86 & $-$0.0005 & 0.0004 & $-$10.72 & $<0.001$ & *** \\
 66 & 2.75 & 2.61 &  93 & $-$0.0005 & 0.0004 & $-$10.65 & $<0.001$ & *** \\
 72 & 3.00 & 2.86 &  85 & $-$0.0005 & 0.0004 & $-$10.23 & $<0.001$ & *** \\
 78 & 3.25 & 2.38 & 102 & $-$0.0005 & 0.0004 & $-$10.22 & $<0.001$ & *** \\
 84 & 3.50 & 2.66 &  92 & $-$0.0005 & 0.0005 & $-$9.28 & $<0.001$ & *** \\
 90 & 3.75 & 2.93 &  83 & $-$0.0004 & 0.0005 & $-$8.28 & $<0.001$ & *** \\
 96 & 4.00 & 2.91 &  84 & $-$0.0005 & 0.0005 & $-$7.74 & $<0.001$ & *** \\
102 & 4.25 & 3.06 &  80 & $-$0.0004 & 0.0005 & $-$6.93 & $<0.001$ & *** \\
108 & 4.50 & 3.08 &  79 & $-$0.0004 & 0.0006 & $-$6.45 & $<0.001$ & *** \\
114 & 4.75 & 3.09 &  79 & $-$0.0004 & 0.0006 & $-$6.23 & $<0.001$ & *** \\
120 & 5.00 & 2.29 & 106 & $-$0.0004 & 0.0007 & $-$7.02 & $<0.001$ & *** \\
126 & 5.25 & 2.60 &  94 & $-$0.0004 & 0.0007 & $-$6.38 & $<0.001$ & *** \\
132 & 5.50 & 2.92 &  84 & $-$0.0005 & 0.0007 & $-$6.40 & $<0.001$ & *** \\
138 & 5.75 & 3.06 &  80 & $-$0.0005 & 0.0007 & $-$6.10 & $<0.001$ & *** \\
144 & 6.00 & 2.68 &  91 & $-$0.0005 & 0.0008 & $-$6.51 & $<0.001$ & *** \\
150 & 6.25 & 3.14 &  78 & $-$0.0005 & 0.0008 & $-$5.43 & $<0.001$ & *** \\
156 & 6.50 & 2.42 & 101 & $-$0.0005 & 0.0008 & $-$5.72 & $<0.001$ & *** \\
162 & 6.75 & 2.49 &  98 & $-$0.0005 & 0.0008 & $-$6.17 & $<0.001$ & *** \\
168 & 7.00 & 2.26 & 108 & $-$0.0005 & 0.0008 & $-$6.90 & $<0.001$ & *** \\
174 & 7.25 & 2.03 & 120 & $-$0.0005 & 0.0009 & $-$6.66 & $<0.001$ & *** \\
180 & 7.50 & 2.14 & 114 & $-$0.0006 & 0.0010 & $-$6.09 & $<0.001$ & *** \\
186 & 7.75 & 2.28 & 107 & $-$0.0005 & 0.0010 & $-$5.48 & $<0.001$ & *** \\
192 & 8.00 & 2.09 & 117 & $-$0.0006 & 0.0011 & $-$5.34 & $<0.001$ & *** \\
198 & 8.25 & 1.75 & 139 & $-$0.0006 & 0.0011 & $-$5.86 & $<0.001$ & *** \\
204 & 8.50 & 1.71 & 142 & $-$0.0006 & 0.0012 & $-$5.97 & $<0.001$ & *** \\
210 & 8.75 & 1.57 & 155 & $-$0.0006 & 0.0012 & $-$6.30 & $<0.001$ & *** \\
216 & 9.00 & 1.63 & 149 & $-$0.0006 & 0.0012 & $-$6.16 & $<0.001$ & *** \\
222 & 9.25 & 1.92 & 127 & $-$0.0006 & 0.0012 & $-$5.66 & $<0.001$ & *** \\
228 & 9.50 & 1.90 & 128 & $-$0.0006 & 0.0013 & $-$5.68 & $<0.001$ & *** \\
234 & 9.75 & 1.88 & 130 & $-$0.0006 & 0.0012 & $-$5.63 & $<0.001$ & *** \\
240 & 10.00 & 1.85 & 132 & $-$0.0006 & 0.0013 & $-$5.55 & $<0.001$ & *** \\
\end{longtable}